\documentclass[11pt]{article}

\usepackage[margin=1in]{geometry}
\usepackage{amsmath}
\usepackage{booktabs}
\usepackage{graphicx}
\usepackage{hyperref}
\usepackage{xcolor}
\newcommand{\method}{Speculative Correction}
\newcommand{\mini}{LLaDA2.1-mini}
\newcommand{\flash}{LLaDA2.1-flash}

\title{Speculative Correction: Draft-then-Refine Decoding for Diffusion Language Models}
\author{
Brian K Chen\thanks{First author. Corresponding author: \texttt{e0694208@u.nus.edu}.}\\
National University of Singapore
\and
Chong Wu\\
City University of Hong Kong
\and
Kenji Kawaguchi \\
National University of Singapore
}
\date{}

\begin{document}
\maketitle

\begin{abstract}
Diffusion language models (DLMs) can revise tokens bidirectionally, but standard decoding procedures often adapt them to left-to-right generation by producing text block by block. We study a simple plug-and-play inference pattern: first generate a complete draft, and then refine the full response with bidirectional diffusion. Using \flash{} and \mini{}, we evaluate two configurations. In flash--flash, the same \flash{} model serves as both drafter and refiner, testing whether an existing model can improve its own block autoregressive output through global refinement. In mini--flash, inspired by speculative decoding, we introduce \emph{speculative correction}: \mini{} drafts a full response, and \flash{} revises it as an editable initialization. Flash--flash improves GSM8K-384 from 0.848 to 0.899 while running $1.20\times$ faster than the selected \flash{} block autoregressive baseline, and improves MBPP-384 from 0.545 to 0.693. Latency-window-matched flash-only controls indicate that these gains persist after targeted tuning of block autoregressive decoding. Causal ablations indicate that completed drafts are useful initializations: refinement from a fully masked span performs poorly, full global refinement provides a clear additional gain on GSM8K, and local refinement captures much of the gain on MBPP and MATH. Mini--flash provides useful quality--latency trade-offs, including MATH-384 performance of 0.294 versus 0.300 for \flash{} while running $2.17\times$ faster. These results support a Pareto-frontier interpretation rather than the claim that the heterogeneous cascade uniformly matches \flash{} quality. Overall, the same-model draft/refine provides evidence that bidirectional refinement is a useful decoding primitive for DLMs, while speculative correction demonstrates a training-free route to fast DLM generation.

\end{abstract}

\section{Introduction}

The meteoric rise in language models in modern society has sparked a frenzied pursuit for more powerful and efficient foundation models. In recent years, diffusion language models (DLMs) \cite{sedd,llada,llada2p1} have garnered increasing interest within the language modeling community. Motivated in part by the success of diffusion models in the vision domain \cite{ddpm,latentdiffusionmodels}, recent work has explored whether diffusion-style denoising can offer advantages that differ from those of autoregressive transformers used in most commercial models such as ChatGPT \cite{openai2024gpt4technicalreport}, Claude \cite{anthropic2024claude3}, DeepSeek \cite{deepseek2024v3}, etc.

Despite showing early promise, especially in terms of efficiency, DLMs face a fundamental challenge when applied to language generation: DLMs are by design inherently bidirectional.  Generation proceeds through denoising over an entire sequence simultaneously, allowing tokens to be revised under full bidirectional context. This makes DLMs architecturally distinct from existing transformers, but also makes them difficult to apply to standard language tasks, which are naturally posed as left-to-right generation problems.  The prevailing solution is \textit{block autoregressive diffusion} (block-AR diffusion): generating text block by block sequentially, where each block is treated with bidirectional attention internally.  This compromise allows DLMs to approximate autoregressive (AR) processes \cite{xue2025anyordergptmaskeddiffusion} to better fit the language modeling paradigm.

\begin{figure}[!t]
\centering
\includegraphics[width=0.90\textwidth]{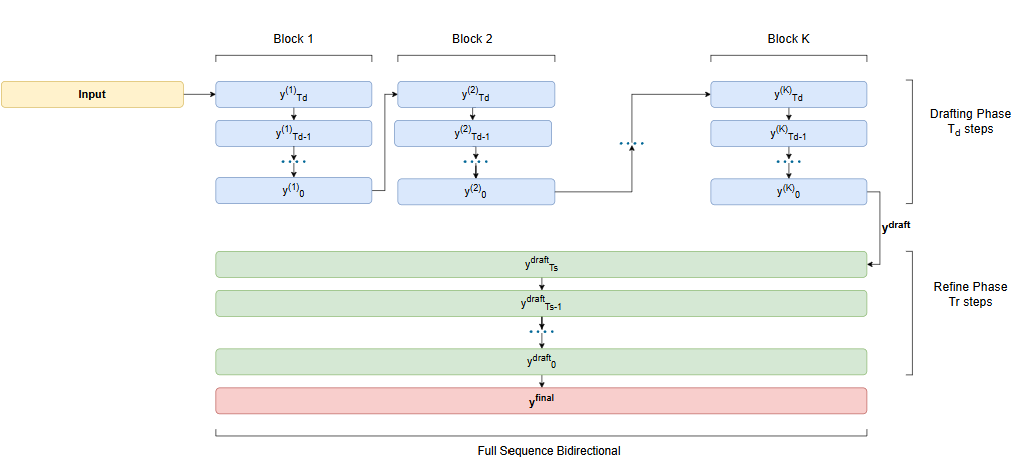}
\caption{Speculative correction. A drafter first produces a complete response, and a bidirectional diffusion refiner then revises the full sequence.}
\label{fig:model-comparison}
\end{figure}

Although practical, this strategy constrains one of the core strengths of diffusion language models. By imposing an approximately autoregressive decoding regime on models designed for bidirectional revision, it limits their ability to revise globally across the sequence. This motivates decoding schemes that use bidirectional refinement more directly, rather than relying primarily on autoregressive approximation.

In this work, we study \emph{draft-then-refine decoding} as a general \textit{plug-and-play} inference pattern for diffusion language models. A first stage rapidly constructs a complete candidate response; a second stage applies full-sequence bidirectional refinement. In this framework, block-AR decoding is used not as a final generation procedure, but as a way to initialize a global corrector. The resulting architecture directly uses the ability of DLMs to review and revise earlier and later tokens jointly.

Success in this setting is not guaranteed. The refiner's denoising process is trained on inputs drawn from its own forward noising distribution, not on outputs of block autoregressive decoding. The draft must provide sufficient structure for the refiner to correct with only a few passes — standard DLM generation from fully masked sequences requires substantially more denoising steps to produce coherent output.

This architecture gives rise to two complementary questions. First, can an existing DLM checkpoint improve its own block-AR generations when computation is reallocated from purely sequential block denoising to complete-sequence drafting followed by global refinement? Second, can a smaller DLM provide a cheap but useful initialization that allows a larger DLM to recover much of its quality at substantially lower cost? We study both questions using the LLaDA2.1 family.

The first setting, \emph{flash--flash}, is the aligned mechanism test: \flash{} serves as both the drafter and the refiner. This configuration is not primarily a speculative-speed system; rather, it isolates whether bidirectional correction is a useful decoding primitive when the draft and refiner are maximally matched. We find positive evidence for this mechanism.  Because no training or architectural modification is involved, any observed gains must originate from capabilities already present in the pretrained checkpoint — an attribution that trained cascade approaches cannot provide. Despite using a computational budget comparable to or lower than the selected strong block-AR \flash{} baseline, flash--flash improves the accuracy on GSM8K-384 from 0.848 to 0.899 while running $1.20\times$ faster and improves the MBPP-384 pass rate from 0.545 to 0.693. A validation-selected latency-window-matched flash-only control further supports that these gains are not simply explained by an undertuned block-AR schedule.

The second setting, \emph{mini--flash}, is the heterogeneous quality--latency setting and the main use case for the method we introduce: \emph{speculative correction}, inspired by speculative decoding. In speculative decoding, a cheap model proposes tokens, and a larger model verifies or rejects them under a left-to-right autoregressive process. In speculative correction, the cheap model instead produces a complete response draft, and the larger model uses bidirectional diffusion to revise that draft as an editable full-response initialization. Thus, the expensive model does not perform token-level acceptance or rejection; it can correct any position using the global context. This resembles cascaded diffusion in vision, where an inexpensive or lower-resolution stage provides a rough initialization that a more capable diffusion stage subsequently refines.

We emphasize that mini--flash is a frontier method, not a claim of uniform dominance over \flash{} block-AR decoding. Its goal is to expose new operating points between mini-only speed and flash-only quality. In this role, the plug-and-play cascade is promising. On MATH-384, mini--flash scores 0.294 against a flash baseline of 0.300 while running $2.17\times$ faster, recovering roughly 79\% of the mini-to-flash quality gap. On MBPP-384, it improves from 0.401 for mini-only decoding to 0.568 after refinement, slightly above the selected flash baseline of 0.545 in observed score while running $1.11\times$ faster. On GSM8K and MATH-512, the cascade remains faster but does not fully recover \flash{} quality. Thus the current evidence supports a quality--latency frontier interpretation rather than an equivalence or dominance claim.

This plug-and-play setting is deliberately stringent. The drafter and the refiner were trained independently: \mini{} was not optimized to produce refinement-friendly drafts, and \flash{} was not trained to interpret the output of \mini{} as partially denoised initializations. In diffusion terms, the drafter's output distribution may not match the refiner's expected denoising trajectory. Therefore, positive frontier points in this setting are especially informative: they suggest that pretrained DLMs already possess nontrivial cross-model correction ability, while also suggesting that drafter training, refiner adaptation, or joint drafter--refiner optimization could further improve the frontier.

To verify that the gains reflect actual correction behavior rather than only aggregate score shifts, we include a matched correctness-transition diagnostic. This is a mechanism check: it asks whether refinement turns incorrect drafts into correct outputs more often than it turns correct outputs into incorrect ones. In the same-model setting, flash--flash fixes more errors than it introduces relative to the selected \flash{} block-AR baseline. In the heterogeneous setting, mini--flash similarly improves over the local \mini{} draft baseline, fixing 111 errors versus 19 introduced errors on GSM8K-384, 84 versus 18 on MATH-384, and 324 versus 66 on MBPP-384. These diagnostics support the central interpretation: refinement often repairs concrete draft errors, while the remaining losses against \flash{} reflect incomplete recovery under plug-and-play model mismatch.

Our work is closely aligned with the concurrent work on Diffusion in Diffusion~\cite{diffusion-in-diffusion}, which also identifies draft-then-refine as a promising structure for block diffusion models. Diffusion in Diffusion realizes this idea through dedicated progressive refinement mechanisms and training. In contrast, we ask whether the same principle can be instantiated directly at inference time using existing pretrained DLMs, without additional training or architectural modification. The same-model setting provides evidence for the decoding primitive itself; the heterogeneous setting suggests that even an unoptimized small-to-large DLM cascade can expose useful quality--latency frontier points.

Our main contributions are:

\begin{itemize}
    \item We formulate draft-then-refine decoding as a general inference architecture for DLMs, using fast complete-sequence drafting followed by global bidirectional correction.

    \item We find that same-model correction can improve existing checkpoints: flash--flash validates the mechanism under maximal alignment and improves over selected strong block-AR \flash{} baselines, including latency-window-matched controls.

    \item We introduce speculative correction, where \mini{} drafts and \flash{} refines. This plug-and-play cascade traces practical quality--latency frontier points.

    \item We analyze causal draft/refine ablations, matched correction transitions, and model-mismatch effects, indicating that completed drafts are useful initializations, global refinement often repairs concrete draft errors, and remaining quality gaps point toward trained or jointly optimized DLM cascades.
\end{itemize}

\section{Related Work}

\paragraph{Diffusion language models.}
Diffusion language models adapt denoising diffusion to token sequences, either by operating over continuous token embeddings or by defining discrete noising processes over categorical states. Early work established diffusion-style text generation through continuous latent or embedding-space denoising \cite{diffusion-lm,diffuseq} and through discrete/categorical diffusion formulations \cite{argmax-flows,d3pm,ctmc-discrete-diffusion,sahoo2024simpleeffectivemaskeddiffusion}. Subsequent work improved the practical quality of discrete and masked diffusion language models, including semi-autoregressive simplex diffusion \cite{ssd-lm}, score-entropy discrete diffusion \cite{sedd}, and large-scale masked diffusion systems such as LLaDA \cite{llada}. LLaDA2.1 further improves the speed--quality trade-off of large diffusion language models through token-to-token editing and releases the \mini{} and \flash{} checkpoints used in this work \cite{llada2p1}. These works establish DLMs as a viable alternative to purely autoregressive generation. Our focus is complementary: rather than proposing a new training objective or a new DLM checkpoint, we study how existing DLMs should be decoded at inference time.

\paragraph{Block and semi-autoregressive diffusion decoding.}
A central practical difficulty for DLMs is that standard language generation is open-ended and prefix-conditioned, whereas fully bidirectional diffusion is most natural for fixed-length sequence completion. Semi-autoregressive and blockwise decoding address this mismatch by generating text in sequential blocks while performing bidirectional denoising inside each block \cite{ssd-lm,block-diffusion}. Block diffusion models interpolate between autoregressive and diffusion language models, enabling flexible-length generation, partial parallelism, and more efficient inference mechanisms such as KV caching \cite{block-diffusion}. This line of work makes DLMs compatible with standard text-completion tasks, but the sequential block boundary still limits global revision: earlier blocks are fixed before later context is available. Our method keeps block diffusion as a fast drafting mechanism, but adds a full-response bidirectional refinement stage so that the model can revise earlier and later tokens jointly.

\paragraph{Draft--then--refine and progressive diffusion.}
The idea of producing a rough sample and then refining it is widespread in diffusion modeling. Cascaded diffusion models for images generate a coarse sample and then apply later diffusion stages, often at higher resolution, to improve fidelity \cite{cascaded-diffusion}. In language diffusion, the closest related work is the concurrent Diffusion in Diffusion paper \cite{diffusion-in-diffusion}. Diffusion in Diffusion identifies a similar draft--then--refine structure for block diffusion models: it drafts with small blocks, selectively remasks low-confidence tokens, and refines with larger or global bidirectional receptive fields, supported by mix-scale training. Our work was developed concurrently and explores the same emerging design principle from a different angle. Rather than training a model specifically for progressive refinement, we ask whether existing pretrained DLM checkpoints already support draft--then--refine decoding as a plug-and-play inference procedure. This distinction is important for both of our settings: same-model correction tests whether one checkpoint can improve its own block-AR output through global refinement, while speculative correction tests whether independently trained small and large DLMs can be composed without joint training.

\paragraph{Speculative decoding for Transformer Models.}
Speculative decoding accelerates autoregressive language models by using a cheap draft model to propose several future tokens and a larger target model to verify those proposals in parallel \cite{speculative-decoding,speculative-sampling}. Modern variants improve the drafter or verification process with additional heads \cite{cai2024medusasimplellminference}, tree-structured proposals \cite{specinfer}, or feature-level predictors \cite{eagle}, but they retain the same basic left-to-right acceptance mechanism: proposed tokens are accepted or rejected according to criteria derived from the target model's autoregressive distribution. \method{} borrows the cost intuition of speculative decoding---use a cheap stage to reduce the work done by an expensive stage---but changes the correction mechanism. The refiner does not accept or reject tokens by token. It receives a complete draft and performs bidirectional diffusion over the full editable response, allowing any position to be revised using global context. Recent work has begun to adapt speculative ideas directly to diffusion language models. DiffuSpec uses a pretrained DLM as a parallel drafter for an autoregressive verifier, adding causal-consistency path search and adaptive draft-length control to make bidirectional DLM drafts compatible with left-to-right verification \cite{diffuspec}. DFlash trains a lightweight block-diffusion drafter conditioned on the features of the target model to improve speculative decoding for autoregressive LLMs \cite{dflash}.

These methods primarily aim to accelerate decoding while preserving, or closely approximating, a target decoding process. Our goal is different: we study draft--then--refine as an editable initialization and correction procedure.

\section{Background: Diffusion Language Models}

We briefly review the diffusion mechanism underlying DLMs to establish the notation and motivate the draft-then-refine approach.

\subsection{Forward and Reverse Processes}

Let $y \in \mathcal{V}^L$ denote a sequence of $L$ tokens from vocabulary $\mathcal{V}$. A diffusion language model defines a forward process that gradually corrupts $y$ into noise, and a reverse process that recovers $y$ from noise.

The forward process begins with the clean sequence $y_0 = y$ and progressively applies noise over $T$ timesteps:
\[
    y_0 \to y_1 \to y_2 \to \cdots \to y_T,
\]
where $y_t$ represents the sequence at noise level $t$. At $t = T$, the sequence approaches pure noise---tokens are essentially uniformly random or masked, depending on the specific DLM formulation.

The reverse process learns to denoise: starting from $y_T$, the model iteratively recovers cleaner versions
\[
    y_T \to y_{T-1} \to \cdots \to y_0.
\]
Each denoising step is parameterized by a neural network $p_\theta(y_{t-1} \mid y_t)$ that predicts the cleaner sequence given the noisier one. Training minimizes the discrepancy between predicted and actual clean tokens across all noise levels.

\subsection{Initialization and Target Distribution}

Crucially, the reverse process requires two components: an \emph{initialization} $y_T$ and a \emph{target distribution} over clean sequences $y_0$. In standard DLM generation:

\begin{itemize}
    \item \textbf{Initialization}: The process starts from pure noise $y_T$, typically sampled uniformly or as a mask token sequence.
    \item \textbf{Target distribution}: The model's training defines an implicit target distribution $p_\theta(y_0)$ over clean sequences. For conditional generation given a prompt $x$, the target is $p_\theta(y_0 \mid x)$.
\end{itemize}

\subsection{Block Autoregressive Diffusion}

Standard fully bidirectional DLM generation is not directly aligned with left-to-right language tasks, since the model denoises a complete sequence. Block autoregressive diffusion provides a practical compromise by partitioning generation into blocks.

Let the response be divided into $K$ blocks: $y = (y^{(1)}, y^{(2)}, \ldots, y^{(K)})$. Block-AR diffusion generates each block sequentially:
\[
    y^{(1)} \to y^{(2)} \to \cdots \to y^{(K)},
\]
where each block $y^{(k)}$ is generated through the full diffusion process with bidirectional attention \emph{within} that block. Critically, when generating block $y^{(k)}$, the model conditions on all previously generated blocks $(y^{(1)}, \ldots, y^{(k-1)})$ as context, but cannot attend to future blocks that have not yet been generated.

Within each block, the model applies $T$ denoising steps:
\[
    y^{(k)}_T \to y^{(k)}_{T-1} \to \cdots \to y^{(k)}_0,
\]
where $y^{(k)}_T$ is initialized as noise and $y^{(k)}_0$ is the clean output for block $k$. Under an equal-cost-per-pass abstraction, the total computational cost for generating a sequence of $K$ blocks with $T$ steps per block is $K \times T$ denoising passes.

\begin{figure}[!t]
\centering
\includegraphics[width=0.88\textwidth]{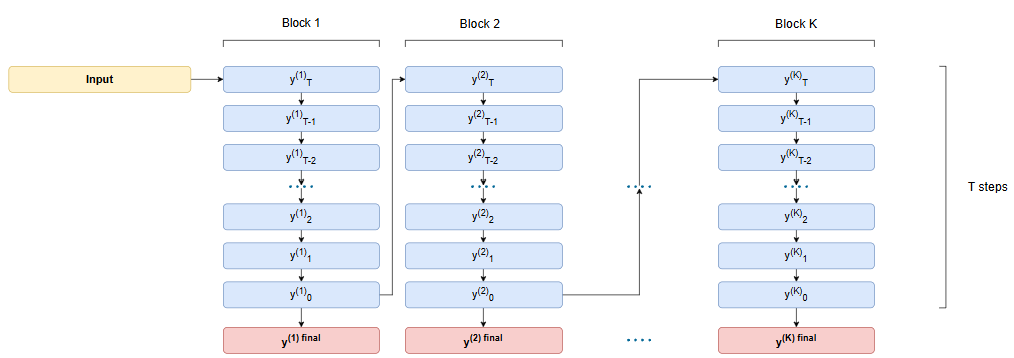}
\caption{Standard block autoregressive diffusion. The model generates response blocks sequentially, preserving compatibility with left-to-right language tasks but limiting later tokens' ability to revise earlier blocks.}
\label{fig:baseline-model}
\end{figure}

\subsection{Connection to Draft-then-Refine}

The draft-then-refine approach exploits a simple observation: if the initialization is already closer to the target distribution, fewer denoising steps may be required. Rather than starting each block from pure noise $y_T$, we can use the drafter's output $y_{\text{draft}}$ as a partially denoised initialization for the refiner.

This perspective reveals why the choice of drafter matters. The drafter produces a sequence from its own target distribution $p_{\theta_s}(y_0 \mid x)$, but the refiner was trained to denoise according to its target distribution $p_{\theta_l}(y_0 \mid x)$. When $\theta_s=\theta_l$ (flash--flash), drafting and refinement use the same model and tokenizer. This removes cross-model mismatch, although block-AR drafts may still differ from the corrupted intermediate states seen during refinement training. When $\theta_s \neq \theta_l$ (mini--flash), the draft distribution may differ more substantially from the refiner's expected inputs, creating a mismatch that the refiner must accommodate.

In image diffusion, cascaded models explicitly account for this by training successive stages to accept outputs from previous stages as initialization. Our plug-and-play approach skips this joint training, which is computationally convenient but introduces distribution mismatch. 

\section{Method}

We consider a prompt $x$ and aim to generate a response $y$. The method proceeds in two stages: drafting and refinement.

\subsection{Drafting}

The drafter produces a complete response using block autoregressive diffusion. Following the notation from Section~3, the response is partitioned into $K$ blocks, and each block is generated sequentially with $T_d$ denoising steps:
\[
    y^{\text{draft}} = D_{\theta_s}(x) = (y^{(1)}, y^{(2)}, \ldots, y^{(K)}).
\]
The drafter can be either the same model as the refiner or a smaller model. In the flash--flash configuration, $D_{\theta_s}$ is \flash{}. In the mini--flash configuration, $D_{\theta_s}$ is \mini{}.

The drafting stage is designed to be computationally inexpensive. It uses fewer denoising steps per block ($T_d < T$) than standard block-AR diffusion, where $T$ is the number of steps required for high-quality generation. The total drafting cost is $K \times T_d$ denoising passes. The draft serves as an initialization for refinement rather than a standalone output.

\subsection{Bidirectional Refinement}

The refiner receives both the prompt and the completed draft. Prompt tokens remain fixed throughout. Draft tokens are subject to revision. Unlike block-AR diffusion, where each block is denoised independently with limited context, the refiner applies bidirectional correction over the \emph{entire} sequence:
\[
    y^{\text{final}} = R_{\theta_l}(x, y^{\text{draft}}).
\]
In all experiments, $R_{\theta_l}$ is \flash{}.

The refinement stage applies $T_r$ denoising steps throughout the sequence. Since the draft $y^{\text{draft}}$ provides a better initialization than pure noise---it already contains a coherent structure and content---the refiner can achieve convergence with fewer steps than would be required when starting from $y_T$. The total refinement cost is $T_r$ denoising passes.

The key difference from block-AR diffusion is that the refiner sees a complete candidate response and can revise any position using context from the entire sequence. Earlier tokens can be corrected based on later tokens, and vice versa. This exploits the bidirectional capability that distinguishes DLMs from autoregressive models.

The total cost of draft-then-refine is $K \times T_d + T_r$ denoising passes. For speedup, we require:
\[
    K \times T_d + T_r < K \times T,
\]
where the right-hand side is the cost of standard block-AR diffusion. Since the draft provides a reasonable approximation, typically $T_d \ll T$ and $T_r$ can also be small, yielding efficiency gains. This equation is only an intuition: mini and flash have different per-pass costs, full-span refinement has a different attention pattern from blockwise decoding, and memory placement can dominate in practice.

\subsection{Model Configurations}

We evaluate two primary configurations.

\paragraph{Flash--flash.}

\flash{} serves as both drafter and refiner. This configuration is not designed to be the most cost-efficient; rather, it tests whether the draft-then-refine mechanism is useful when the two stages are maximally aligned. The drafter and the refiner share the same tokenizer and parameters, and they differ primarily in the decoding schedule and the role that the generated text plays. This setting removes cross-model mismatch as a confound and tests whether reallocating inference compute into complete-sequence drafting followed by global refinement can improve over block-AR decoding.

\paragraph{Mini--flash (Speculative Correction).}

\mini{} drafts and \flash{} refines. This configuration tests whether a cheaper draft can reduce computational cost while allowing a larger model to recover quality. The models share a tokenizer and architectural family, which eases alignment, but they were trained independently. This represents the speculative correction setting, analogous to speculative decoding for autoregressive models but using bidirectional refinement rather than token-level verification.

\section{Experimental Setup}

\subsection{Models}

All experiments use \mini{} and \flash{} from the LLaDA2.1 family~\cite{llada2p1}. We use publicly available checkpoints directly without any training, distillation, or fine-tuning. The models share a tokenizer and architectural family, which facilitates plug-and-play composition.

Our choice of LLaDA2.1 is driven by the requirements of the proposed decoding procedure rather than by a claim that it is uniquely suitable among DLM families. The draft--then--refine procedure requires a model family that can efficiently produce complete drafts with block-autoregressive diffusion, but also supports bidirectional refinement over an editable generated span while keeping the prompt fixed. The heterogeneous setting further requires at least two publicly available checkpoints from the same family, so that the drafter and refiner share a tokenizer, architectural design, and broadly compatible training distribution. LLaDA2.1 satisfies this combination of requirements for our experiments: it provides both \mini{} and \flash{}, supports practical block-AR generation, and exposes an editing-based diffusion mechanism compatible with full-response refinement. Other recent open block-diffusion families, such as SDAR \cite{sdar} and Fast-dLLM v2 \cite{fastdllm}, provide valuable multi-size block-diffusion systems, but their public decoding formulations are primarily autoregressive over blocks with bidirectional denoising within the current block. Under their public interfaces, they therefore do not directly instantiate the full-response bidirectional refiner studied here without modifying the sampler or reducing our method to a block-local refinement variant. We consequently use LLaDA2.1 as the main testbed and view extension to future multi-size, fully editable DLM families as an important direction for follow-up work.

\subsection{Tasks}

We primarily evaluate on GSM8K, MATH, MBPP, HumanEval, and synthetic arithmetic.  These tasks require multi-step reasoning or structured generation, settings where bidirectional revision may be particularly beneficial. 

The suffixes 384 and 512 denote maximum generated response length, excluding prompt tokens. The same value is also the default full-span refinement budget in cascade runs. GSM8K uses the test split with the prompt ``Solve the following grade-school math problem'' and asks for a final numeric answer. The fixed MATH results use 500 examples from the MATH-500 evaluation subset and ask the model to end with ``Final answer: \textit{answer}''. MBPP uses the sanitized test split and standard assertion tests. HumanEval uses all 164 standard tasks. Synthetic-expanded and synthetic-varied are generated arithmetic word-problem sets with known numeric answers; expanded uses many GSM8K-style templates, while varied adds receipt, roster, manifest, email, table, and memo-style formats to reduce template homogeneity.

The fixed-configuration runs use the completed run batches shown in Appendix~\ref{app:stats}: GSM8K has 1115 unique IDs at the 384-token budget and 793 at the 512-token budget; MATH has 500 unique IDs; MBPP has 257; HumanEval has all 164; and each synthetic set has 1000. The repeated-seed structure is handled by the problem-level confidence intervals described below. Invalid generations are not filtered out: missing numeric answers, syntax errors, failed assertions, and timeouts are counted as incorrect.

\subsection{Baselines and Metrics}

We compare against standard block-autoregressive decoding with both \flash{} and
\mini{}. The \flash{} baseline is the main target because it is the stronger
single-stage model that draft--then--refine aims to improve or accelerate.
The \mini{} baseline is reported only for heterogeneous mini--flash runs, where
it measures how much of the small-to-large quality gap is recovered.

Configurations are chosen in two stages. We first run exploratory sweeps over
block length, denoising steps, thresholds, draft steps, and refinement steps to
identify promising score--latency regions. We then freeze representative
operating points and rerun them at larger scale. Main confidence intervals are
computed only on these fixed runs and should be interpreted as uncertainty
estimates for the selected configurations, not as post-selection-adjusted
guarantees over the full sweep. The exploratory process was applied to both
draft--then--refine configurations and flash-only baselines, so the comparison
is not between a tuned proposed method and an untuned default baseline.
Nevertheless, this does not remove all post-selection bias, since the cascade
family has additional degrees of freedom. A fully locked validation/test
protocol over the entire search space would require substantially more compute
than was available for this study. Configuration details, sweep summaries, and
statistical-unit details are given in the appendix.

To address the possibility that the flash--flash gains come from comparing
against an undertuned flash-only decoder, we additionally run a
latency-window-matched flash-only control. For each selected flash--flash
cascade, we sweep flash-only block-AR decoding over block size, denoising steps,
confidence threshold, and maximum generation length on a validation subset. We
then select the highest-scoring flash-only configuration whose validation
latency is within $\pm 15\%$ of the corresponding cascade latency. In the final
comparison, we evaluate on matching problem IDs and compare flash--flash against
the stronger of the original fixed flash baseline and the validation-selected
flash-only baseline. This prevents the latency-window control from becoming a
weaker comparator than a flash baseline already reported elsewhere.

Finally, we run a causal draft/refine ablation on three representative
384-token settings: GSM8K flash--flash, MBPP flash--flash, and MATH
mini--flash. These runs use the same fixed problem IDs as the corresponding
main tables. For each setting we compare the selected block-AR \flash{}
baseline, the exact draft before refinement, mask-only refinement with the
same refinement-stage budget but no draft pass, local 64-token refinement of
the draft, and the full global draft--then--refine method. This isolates whether the
draft is useful as an initialization and whether the global full-sequence
correction step contributes beyond additional local denoising.

For GSM8K, MATH, and synthetic arithmetic, we report final-answer accuracy after
task-specific extraction and normalization. For MBPP and HumanEval, we report
pass@1 execution success under a fixed code-execution harness. Extraction,
normalization, timeout, and execution details are provided in the appendix.

Efficiency is measured as mean model seconds per example. For baselines this is
the timed block-AR generation call; for cascades it is the sum of the timed draft
and refinement calls. Speedup is measured relative to the selected \flash{}
baseline, so values above $1$ indicate that the method is faster than \flash{}.
Timings are within-environment comparisons using the same evaluation loop,
hardware class, precision, decoding implementation, and metrics.

For mini--flash, we also report the fraction of the \mini{}--to--\flash{}
quality gap recovered,
\[
\frac{\text{Score}_{\text{mini--flash}}-\text{Score}_{\text{mini}}}
     {\text{Score}_{\text{flash}}-\text{Score}_{\text{mini}}}.
\]
A value of $0$ matches \mini{}, $1$ reaches the selected \flash{} baseline, and
values above $1$ exceed the observed \flash{} score. Because this ratio can be
unstable when the mini--flash gap is small, we interpret it alongside absolute
scores, delta columns, and confidence intervals.

\section{Results}

\subsection{Same-Model Correction Improves Existing Checkpoints}

Table~\ref{tab:flash-fixed} evaluates same-model correction, where \flash{} is used as both drafter and refiner. This is not merely an ablation: it tests whether an existing checkpoint can improve itself at inference time by reallocating computation from purely blockwise decoding to complete-sequence drafting followed by global bidirectional refinement. ``IDs'' is the number of unique problem IDs after grouping repeated seeds. The confidence interval is a bootstrap over problem-level paired means for $\Delta_{\text{vs flash}}$.

\begin{table}[t]
\centering
\small
\resizebox{\textwidth}{!}{
\begin{tabular}{lrrrrrrrrr}
\toprule
Task & Len & IDs & Flash & Flash--flash & $\Delta_{\text{vs flash}}$ & 95\% CI & Flash s/ex & Method s/ex & Speedup \\
\midrule
GSM8K & 384 & 1115 & 0.848 & 0.899 & +0.051 & [+0.028, +0.075] & 19.38 & 16.09 & 1.20$\times$ \\
GSM8K & 512 & 793 & 0.850 & 0.907 & +0.057 & [+0.029, +0.086] & 19.91 & 16.39 & 1.22$\times$ \\
HumanEval & 384 & 164 & 0.415 & 0.622 & +0.207 & [+0.140, +0.280] & 7.94 & 9.16 & 0.87$\times$ \\
HumanEval & 512 & 164 & 0.415 & 0.622 & +0.207 & [+0.140, +0.274] & 8.01 & 8.83 & 0.91$\times$ \\
MBPP & 384 & 257 & 0.545 & 0.693 & +0.148 & [+0.082, +0.214] & 7.34 & 8.10 & 0.91$\times$ \\
MBPP & 512 & 257 & 0.549 & 0.696 & +0.148 & [+0.082, +0.218] & 7.62 & 8.31 & 0.92$\times$ \\
\bottomrule
\end{tabular}}
\caption{Same-model correction results. $\Delta_{\text{vs flash}}$ is flash--flash minus the selected \flash{} block-diffusion baseline. Scores and 95\% confidence intervals average repeated generations per problem before bootstrapping over problem IDs. Timing is mean model seconds per evaluated example.}
\label{tab:flash-fixed}
\end{table}

Same-model correction provides the cleanest evidence that draft-then-refine is a useful decoding mechanism for DLMs. On GSM8K-384, flash--flash improves accuracy from 0.848 to 0.899 while running $1.20\times$ faster than the selected flash baseline; on GSM8K-512, it improves from 0.850 to 0.907 while running $1.22\times$ faster. The problem-level confidence intervals exclude zero in both cases. On MBPP, flash--flash improves pass rate by +0.148 at both token budgets, and HumanEval shows a similar quality gain from 0.415 to 0.622. The code results are slower than the selected flash baseline, but they show that same-model global refinement can substantially improve quality when drafter and refiner are aligned.

\subsection{Latency-Window-Matched Flash-Only Controls}

Table~\ref{tab:latency-window-flash-control} reports the latency-window-matched
flash-only control. For each flash--flash setting, we first selected a
flash-only block-AR configuration on a validation subset, constrained to lie
within a $\pm 15\%$ latency window around the corresponding cascade. We then
formed a conservative final comparator: on the final matched problem IDs, we
take the stronger of the original fixed flash baseline and the
validation-selected flash-only baseline. This experiment tests whether the
flash--flash gains persist after giving block-AR decoding its own targeted
latency-window sweep.

\begin{table}[t]
\centering
\small
\resizebox{\textwidth}{!}{
\begin{tabular}{llrrrrrrr}
\toprule
Task & IDs & Fixed flash (matched IDs) & Val.-selected flash & Comparator & Flash--flash & $\Delta$ & 95\% CI & Speedup \\
\midrule
GSM8K-384 & 500 & 0.832 / 19.08s & 0.832 / 19.45s & 0.832 / 19.08s & 0.900 / 15.61s & +0.068 & [+0.032, +0.104] & 1.22$\times$ \\
GSM8K-512 & 500 & 0.846 / 19.46s & 0.832 / 19.26s & 0.846 / 19.46s & 0.918 / 15.80s & +0.072 & [+0.036, +0.106] & 1.23$\times$ \\
HumanEval-384 & 164 & 0.415 / 7.94s & 0.561 / 9.70s & 0.561 / 9.70s & 0.622 / 9.16s & +0.061 & [-0.030, +0.159] & 1.06$\times$ \\
HumanEval-512 & 164 & 0.415 / 8.01s & 0.561 / 10.12s & 0.561 / 10.12s & 0.622 / 8.83s & +0.061 & [-0.030, +0.152] & 1.15$\times$ \\
MBPP-384 & 257 & 0.545 / 7.34s & 0.510 / 7.61s & 0.545 / 7.34s & 0.693 / 8.15s & +0.148 & [+0.082, +0.218] & 0.90$\times$ \\
MBPP-512 & 257 & 0.549 / 7.62s & 0.510 / 7.65s & 0.549 / 7.62s & 0.696 / 8.31s & +0.148 & [+0.082, +0.218] & 0.92$\times$ \\
\bottomrule
\end{tabular}}
\caption{Latency-window-matched flash-only control on matching problem IDs. The validation-selected flash baseline is selected from a sweep over block size, denoising steps, confidence threshold, and maximum generation length with a $\pm 15\%$ validation-latency window. The final comparator is the stronger of the original fixed flash baseline and the validation-selected flash baseline on the same IDs. $\Delta$ is flash--flash minus the comparator, with a problem-level bootstrap confidence interval. Speedup is comparator seconds divided by flash--flash seconds, so values above $1$ mean flash--flash is faster.}
\label{tab:latency-window-flash-control}
\end{table}

The latency-window-matched control strengthens the same-model conclusion. On
GSM8K, the original fixed flash baseline is at least as strong as the
validation-selected flash baseline on the matched final IDs, so it becomes the
comparator. Flash--flash still improves by +0.068 at 384 tokens and +0.072 at
512 tokens, with paired intervals that exclude zero, while also being faster.
On MBPP, the original flash baseline is also the stronger comparator; even
against it, flash--flash improves pass rate by +0.148 at both token budgets.
HumanEval is the only case where validation tuning produces the stronger
flash-only comparator. The observed gain remains +0.061, but the paired
interval crosses zero, so we treat HumanEval as supportive but not conclusive.
Thus, the main flash--flash result is not explained by a weak or obviously
undertuned flash-only block-AR comparator.

\subsection{Causal Draft/Refine Ablation}
\label{sec:causal-ablation}

The latency-window control tests whether a better tuned block-AR schedule can
explain the flash--flash gains. We next ask a more mechanistic question: which
part of draft--then--refine matters? Table~\ref{tab:causal-ablation} evaluates
four ablations on the same fixed IDs as the main tables. \textit{Exact draft
only} scores the completed draft before refinement. \textit{Mask-only
refinement} uses the same refinement-stage budget as the proposed method but
omits the draft pass and starts from a fully masked generated span. This row
tests whether the completed draft is a useful initialization; it is not a
compute-matched all-mask decoder with equal total compute. \textit{Local
refinement} starts from the same draft but refines 64-token chunks locally
instead of revising the whole generated span at once. The last row is the
proposed full global draft--then--refine method.

\begin{table}[t]
\centering
\small
\resizebox{\textwidth}{!}{
\begin{tabular}{llrrrrr}
\toprule
Setting & Variant & Score & s/ex & Full $-$ variant & 95\% CI & Fixed / harmed \\
\midrule
GSM8K-384 flash--flash & Block-AR flash & 0.848 & 18.77 & +0.051 & [+0.028, +0.074] & 116 / 59 \\
GSM8K-384 flash--flash & Exact draft only & 0.677 & 14.37 & +0.222 & [+0.196, +0.247] & 248 / 1 \\
GSM8K-384 flash--flash & Mask-only refinement & 0.004 & 1.32 & +0.894 & [+0.876, +0.912] & 997 / 0 \\
GSM8K-384 flash--flash & Local 64-token refine & 0.872 & 17.58 & +0.027 & [+0.015, +0.039] & 37 / 7 \\
GSM8K-384 flash--flash & Full global draft--refine & 0.899 & 15.96 & -- & -- & -- \\
\midrule
MBPP-384 flash--flash & Block-AR flash & 0.545 & 7.14 & +0.148 & [+0.082, +0.214] & 60 / 22 \\
MBPP-384 flash--flash & Exact draft only & 0.510 & 6.95 & +0.183 & [+0.132, +0.233] & 50 / 3 \\
MBPP-384 flash--flash & Mask-only refinement & 0.000 & 0.92 & +0.693 & [+0.634, +0.747] & 178 / 0 \\
MBPP-384 flash--flash & Local 64-token refine & 0.681 & 8.57 & +0.012 & [$-$0.012, +0.035] & 6 / 3 \\
MBPP-384 flash--flash & Full global draft--refine & 0.693 & 8.01 & -- & -- & -- \\
\midrule
MATH-384 mini--flash & Block-AR flash & 0.300 & 43.31 & $-$0.006 & [$-$0.044, +0.034] & 48 / 51 \\
MATH-384 mini--flash & Exact draft only & 0.272 & 16.91 & +0.022 & [+0.008, +0.038] & 14 / 3 \\
MATH-384 mini--flash & Mask-only refinement & 0.042 & 1.87 & +0.252 & [+0.212, +0.294] & 135 / 9 \\
MATH-384 mini--flash & Local 64-token refine & 0.296 & 22.92 & $-$0.002 & [$-$0.012, +0.008] & 3 / 4 \\
MATH-384 mini--flash & Full global draft--refine & 0.294 & 19.19 & -- & -- & -- \\
\bottomrule
\end{tabular}}
\caption{Causal draft/refine ablation on fixed problem IDs. ``Full $-$ variant'' is the problem-level paired difference between full global draft--then--refine and the ablated variant, with a paired bootstrap confidence interval. Mask-only refinement uses the same refinement-stage budget as the proposed method but omits the draft pass; it tests whether the completed draft is a useful initialization, not whether an all-mask decoder with equal total compute could match the cascade. ``Fixed / harmed'' counts problem IDs where full global refinement changes an incorrect ablation output to correct, or a correct ablation output to incorrect.}
\label{tab:causal-ablation}
\end{table}

The ablation supports the central mechanism while also showing its limits. On GSM8K-384 flash--flash, the completed draft alone scores only 0.677, while full global refinement reaches 0.899; mask-only refinement with the same refinement-stage budget scores 0.004 — near-total failure. Even with 52 edit passes (Appendix C.4), mask-only refinement reaches only 0.009. While DLMs are in principle capable of generating from fully masked sequences, LLaDA2.1's denoising process either requires substantially more steps than budgeted here or is insufficiently trained for open-ended generation at this sequence length. Either way, the draft provides structure that the refiner cannot efficiently recover on its own, confirming that the two stages play complementary roles. Local refinement is much stronger than the draft, reaching 0.872, but full global refinement still adds +0.027 with a confidence interval that excludes zero. On MBPP-384, full global refinement again beats the draft-only, mask-only, and block-AR rows, but the gap over local refinement is small and the interval crosses zero. In MATH-384 mini--flash, refinement improves significantly over the mini draft and over mask-only refinement, but local and global refinements are statistically tied and neither clearly beats the \flash{} block-AR baseline. Thus, the causal picture is task dependent: complete drafts are consistently useful, global full-sequence correction is decisive on GSM8K, and the heterogeneous MATH setting remains better interpreted as a fast frontier point than as evidence of a universal global-refinement advantage.

\subsection{Speculative Correction Traces Frontier Points}

Table~\ref{tab:miniflash-fixed} evaluates speculative correction, where \mini{} drafts and \flash{} refines. These runs test whether a cheaper untrained drafter can produce useful operating points between mini-only speed and flash-only quality. Here, the mini baseline is conceptually important: it shows whether the cascade actually improves over the small-model control rather than merely inheriting its errors.

\begin{table}[t]
\centering
\small
\resizebox{\textwidth}{!}{
\begin{tabular}{lrrrrrrrrrrr}
\toprule
Task & Len & IDs & Mini & Flash & Mini--flash & $\Delta_{\text{mini}}$ & $\Delta_{\text{flash}}$ & 95\% CI & Gap rec. & Times m/f/c & Speedup \\
\midrule
GSM8K & 384 & 1115 & 0.703 & 0.848 & 0.752 & +0.048 & $-$0.096 & [$-$0.126, $-$0.066] & 34\% & 12.85 / 19.53 / 14.27 & 1.37$\times$ \\
GSM8K & 512 & 793 & 0.695 & 0.850 & 0.741 & +0.047 & $-$0.108 & [$-$0.144, $-$0.074] & 30\% & 15.43 / 20.07 / 16.66 & 1.20$\times$ \\
MATH & 384 & 500 & 0.272 & 0.300 & 0.294 & +0.022 & $-$0.006 & [$-$0.046, +0.032] & 79\% & 16.72 / 40.90 / 18.82 & 2.17$\times$ \\
MATH & 512 & 500 & 0.314 & 0.392 & 0.336 & +0.022 & $-$0.056 & [$-$0.096, $-$0.016] & 28\% & 23.92 / 49.60 / 25.85 & 1.92$\times$ \\
MBPP & 384 & 257 & 0.401 & 0.545 & 0.568 & +0.167 & +0.023 & [$-$0.054, +0.101] & 116\% & 5.42 / 7.46 / 6.73 & 1.11$\times$ \\
MBPP & 512 & 257 & 0.401 & 0.549 & 0.564 & +0.163 & +0.016 & [$-$0.062, +0.089] & 110\% & 6.75 / 7.71 / 7.99 & 0.97$\times$ \\
Synth-exp. & 384 & 1000 & 0.661 & 0.850 & 0.740 & +0.079 & $-$0.110 & [$-$0.142, $-$0.079] & 42\% & 7.13 / 16.22 / 8.58 & 1.89$\times$ \\
Synth-exp. & 512 & 1000 & 0.661 & 0.850 & 0.740 & +0.079 & $-$0.110 & [$-$0.141, $-$0.079] & 42\% & 7.49 / 16.48 / 8.75 & 1.88$\times$ \\
Synth-var. & 384 & 1000 & 0.664 & 0.797 & 0.769 & +0.105 & $-$0.028 & [$-$0.062, +0.005] & 79\% & 5.96 / 11.13 / 7.49 & 1.49$\times$ \\
Synth-var. & 512 & 1000 & 0.664 & 0.797 & 0.769 & +0.105 & $-$0.028 & [$-$0.060, +0.005] & 79\% & 5.74 / 10.32 / 7.05 & 1.46$\times$ \\
\bottomrule
\end{tabular}}
\caption{Speculative correction results. $\Delta_{\text{flash}}$ is mini--flash minus the selected \flash{} baseline; $\Delta_{\text{mini}}$ is mini--flash minus the local \mini{} baseline. Scores and intervals use problem-level means. ``Gap rec.'' is the fraction of the mini-to-flash quality gap recovered. ``Times m/f/c'' gives mean model seconds/example for mini, flash, and cascade.}
\label{tab:miniflash-fixed}
\end{table}

\begin{figure}[t]
\centering
\includegraphics[width=\textwidth]{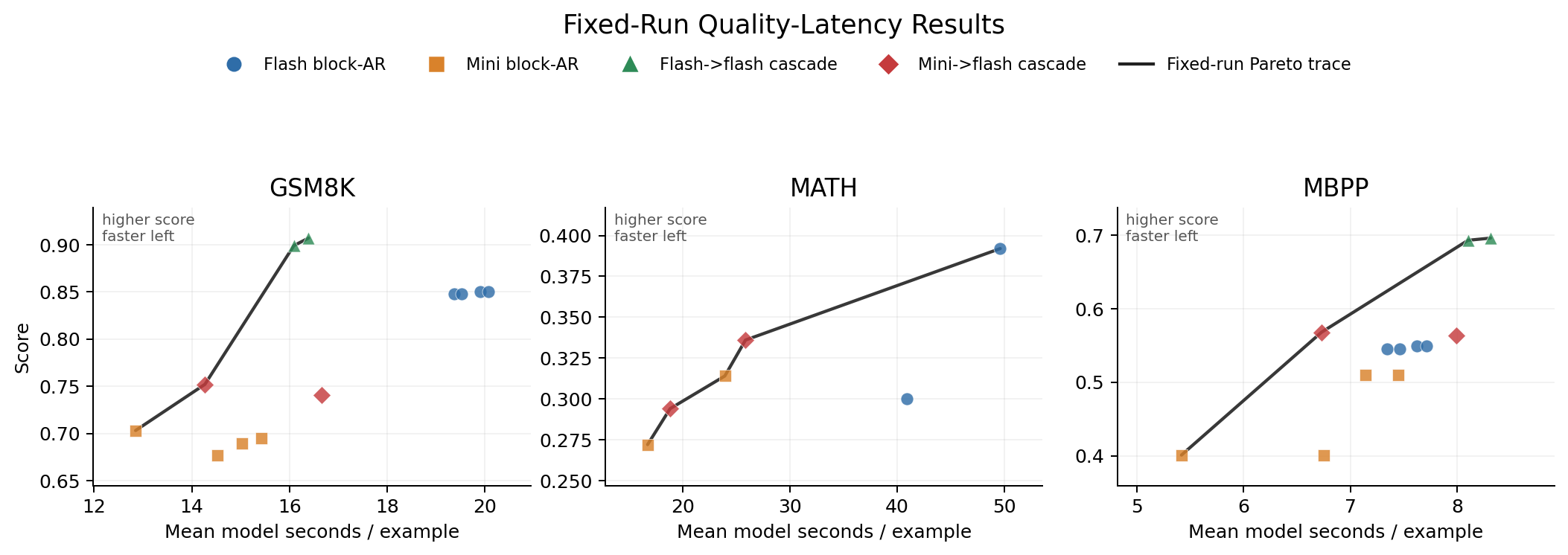}
\caption{Fixed-run quality--latency results for GSM8K, MATH, and MBPP. The plot uses only the completed fixed-configuration rows from the extended confidence-interval archive, excluding small exploratory runs. The x-axis is mean model seconds per example, so better points move up and to the left. The black curve traces the nondominated envelope within each task panel.}
\label{fig:pareto-frontiers}
\end{figure}

Speculative correction behaves differently. As shown in Figure~\ref{fig:pareto-frontiers}, mini--flash is not a universal replacement for flash-only decoding; instead, it defines intermediate quality-cost operating points and traces the Pareto frontier. The strongest frontier result is MATH-384: mini--flash scores 0.294 against a flash baseline of 0.300, with a problem-level interval of [$-$0.046, +0.032], while running $2.17\times$ faster. The interval allows a modest degradation, so this should not be interpreted as proof of equivalence, but the observed point is within 0.006 of flash at less than half the model time. MBPP-384 is also favorable: mini--flash improves over mini from 0.401 to 0.568, slightly exceeds the selected flash baseline of 0.545 in observed score, and runs $1.11\times$ faster, although the confidence interval against flash crosses zero.

The gap-recovery column makes the frontier behavior clearer. GSM8K mini--flash loses measurable accuracy against flash, but it still recovers roughly one-third of the mini-to-flash quality gap while running faster than flash. MATH-384 and synthetic-varied recover about 79\% of the gap with large speedups. MBPP recovers more than the observed mini-to-flash gap, although MBPP-512 is slightly slower than flash. The synthetic arithmetic runs reproduce the same pattern in less memorized arithmetic: the cascade is much faster than flash and better than mini, but below the flash baseline in the expanded setting. HumanEval mini--flash is not a Pareto point because it is slower than flash and does not produce a reliable quality gain; we report it as a negative control in Appendix~\ref{app:negative-controls}.

Some 384- and 512-token rows have identical scores. This is expected for tasks whose extracted final answers or pass/fail outcomes already fit within the shorter budget. The longer budget can still change latency and raw text, but if the extracted answer or executed solution is unchanged, the reported score is unchanged. 

\subsection{Correction Transitions}

The aggregate scores above can obscure what the refinement stage is doing. Table~\ref{tab:correction-transitions} therefore reports a direct transition diagnostic. ``Fixed'' counts cases where the reference system was wrong and the cascade was correct; ``Introduced'' counts cases where the reference was correct and the cascade was wrong.

This table is intentionally a raw matched-generation diagnostic. ``IDs'' counts distinct underlying benchmark examples, while ``Pairs'' counts matched generated examples after repeated seeds are included. For MBPP, for example, 257 problems evaluated across six seeds produce 1542 matched pairs. The fixed and introduced counts are therefore pair-level counts; the main score uncertainty remains the problem-level bootstrap reported in Tables~\ref{tab:flash-fixed} and~\ref{tab:miniflash-fixed}.

\begin{table}[t]
\centering
\small
\resizebox{\textwidth}{!}{
\begin{tabular}{lllrrrrrr}
\toprule
Task & Setting & Reference & IDs & Pairs & Fixed & Introduced & Net & Fixed rate \\
\midrule
GSM8K-384 & flash--flash & flash & 1115 & 2000 & 217 & 101 & +116 & 10.8\% \\
GSM8K-512 & flash--flash & flash & 793 & 1000 & 108 & 48 & +60 & 10.8\% \\
MBPP-384 & flash--flash & flash & 257 & 1542 & 360 & 132 & +228 & 23.3\% \\
GSM8K-384 & mini--flash & mini & 1115 & 2000 & 111 & 19 & +92 & 5.6\% \\
MATH-384 & mini--flash & mini & 500 & 3000 & 84 & 18 & +66 & 2.8\% \\
MBPP-384 & mini--flash & mini & 257 & 1542 & 324 & 66 & +258 & 21.0\% \\
Synth-varied-384 & mini--flash & mini & 1000 & 1000 & 139 & 34 & +105 & 13.9\% \\
\bottomrule
\end{tabular}}
\caption{Pair-level correction transition diagnostic from saved matched generations. For flash--flash, the reference is the selected \flash{} block-AR baseline. For mini--flash, the reference is the local \mini{} baseline. ``Pairs'' counts matched generated examples.}
\label{tab:correction-transitions}
\end{table}

This diagnostic supports the interpretation that refinement is not just changing the aggregate score distribution. In the same-model setting, flash--flash produces a large positive net correction relative to the flash block-AR baseline. In the heterogeneous setting, mini--flash produces positive net correction relative to the mini baseline, which is the relevant draft-quality control. The remaining negative comparisons against flash on GSM8K and synthetic-expanded arithmetic are therefore best interpreted as incomplete recovery of the stronger model's quality, rather than as a failure to improve the draft.

\section{Discussion}
The flash--flash and mini--flash configurations provide complementary evidence about the draft-then-refine architecture. Flash--flash, where the same model serves as both drafter and refiner, removes cross-model mismatch and isolates the effect of the decoding structure. The observed gains---+0.051 with $1.20\times$ speedup on GSM8K-384, +0.057 with $1.22\times$ speedup on GSM8K-512, and +0.148 on MBPP---indicate that the two-stage decomposition can enable more effective computation allocation than the selected block-AR baselines when stages are aligned. The latency-window-matched flash-only controls strengthen this interpretation: after validation tuning over flash-only block size, steps, threshold, and generation length, flash--flash still beats the stronger of the original and validation-selected flash-only baselines on matched final IDs.

The causal ablation helps clarify what the two stages contribute. Exact drafts are not sufficient on their own, and mask-only refinement with the same small refinement-stage budget is much weaker than refinement from a completed draft. This supports the initialization view: the draft gives the refiner a useful point near the target distribution. The mask-only row is intentionally not total-compute matched; it removes the draft to test the value of initialization. The evidence for global full-sequence refinement is strongest on GSM8K, where full global refinement significantly beats local 64-token refinement. On MBPP and MATH, local refinement captures most of the gain, and the full-vs-local intervals cross zero or are very small. Thus, the most conservative reading is that complete-sequence drafting is essential, while the marginal value of global rather than local refinement is task dependent.

Mini--flash extends this architecture to the heterogeneous setting, where a smaller model drafts and a larger model refines. The results are mixed, but informative. On MATH-384, mini--flash is close to flash-only decoding and the problem-level interval includes zero, while the cascade runs $2.17\times$ faster. This shows a strong frontier point, but not formal equivalence: the interval still allows modest degradation. On MBPP-384, it provides a positive frontier point: +0.023 over flash, +0.167 over mini, and $1.11\times$ speedup over flash, although the confidence interval against flash crosses zero. However, on GSM8K and MATH-512, the cascade incurs accuracy penalties while remaining faster than flash-only generation. The task-dependent variation in mini--flash performance may reflect differences in how well the drafter's output aligns with the refiner's target distribution across domains, though further investigation would be required to isolate this mechanism from other factors.

The plug-and-play approach has further limitations. The drafter was trained as a general-purpose model, not to produce refinement-friendly outputs---objectives such as accurate length prediction, structural fidelity, and uncertainty representation may be more important than raw generation quality for this role. The refiner is also trained under its own noising and denoising process, not explicitly on samples from another model used as editable initializations. These mismatches likely contribute to the quality penalties observed in some mini--flash configurations, and suggest that jointly trained drafter-refiner systems could achieve stronger results. 

The evaluation protocol is also limited by compute. We sweep both cascade and
flash-only decoding to avoid comparing a tuned method against a lazy baseline,
and we add a latency-window-matched flash-only control to test baseline
undertuning directly. However, we do not claim that the reported intervals are
locked held-out estimates over the full hyperparameter search. They are
uncertainty estimates for selected operating points, with additional robustness
checks against obvious baseline undertuning.

\section{Future Work}
The current findings point to several directions for improvement. A model trained specifically for the drafter role could be optimized to produce outputs that serve as effective initializations for refinement. Relevant objectives include accurate length prediction, preservation of global structure, and explicit representation of uncertainty. Even without modifying the refiner, fine-tuning the drafter for these properties could improve alignment and reduce the quality penalties observed in some mini--flash configurations.

Furthermore, training the drafter and refiner together would allow separate models to adapt to their roles. The drafter could learn to produce outputs that the refiner can efficiently correct, while the refiner could learn to interpret draft outputs as partial denoisings. This could directly reduce the distribution mismatch that the plug-and-play approach cannot eliminate.

\section{Conclusion}

Diffusion language models should be decoded in a way that exploits their defining capability: bidirectional revision. Block autoregressive diffusion, while practical, forces DLMs to operate against their architectural nature. We have studied a simple alternative that works with the architecture rather than against it: draft a complete sequence, then refine it bidirectionally.

The results support two main conclusions. First, draft-then-refine is intrinsically useful: flash--flash improves quality on MBPP and HumanEval and improves both quality and speed on GSM8K under the selected block-AR baselines, and these gains persist against validation-selected latency-window-matched flash-only controls. The causal ablation further shows that the draft is a useful initialization and that full global refinement can add value beyond local refinement, most clearly on GSM8K. Second, speculative correction traces practical Pareto frontiers: mini--flash is within 0.006 of flash on MATH-384 while running $2.17\times$ faster, and it improves substantially over mini on all main extended comparisons.

We view this work as establishing the drafter-refiner architecture as a viable design pattern for DLMs and providing an early proof-of-concept for the speculative correction methodology. Existing models already produce useful frontier points, but the mixed mini--flash results also make clear that plug-and-play composition is only a starting point. The next step is training: drafters should be optimized not only for accuracy but for producing refinable outputs.

\bibliography{references}

\appendix

The appendix is organized from reproducibility details and statistical interpretation to supplemental fixed-run evidence and exploratory sweeps.

\section{Evaluation and Data Details}
\label{app:evaluation-details}

All benchmark examples are evaluated without filtering invalid outputs. GSM8K uses the \texttt{gsm8k/main} test split and extracts the answer after the dataset's \texttt{\#\#\#\#} marker. MATH uses \texttt{HuggingFaceH4/MATH-500} with the test split. MBPP uses the sanitized split from the Google Research MBPP dataset, falling back to the \texttt{mbpp} dataset name if needed by the local datasets installation. HumanEval uses \texttt{openai\_humaneval}, falling back to \texttt{openai/openai\_humaneval}. Prompt templates and dataset loading are shared across all compared systems.

Numeric tasks use the same scoring implementation across all systems. GSM8K-style tasks compare the last extractable number, including fractions and comma-formatted numbers, with tolerance $10^{-6}$. MATH-style tasks first prefer the last \texttt{\textbackslash boxed\{\}} or \texttt{\textbackslash fbox\{\}} expression, then an explicit final-answer marker, then an ``answer is'' phrase, and finally the last extractable number. If numeric comparison fails, normalized symbolic string matching is used as a fallback.

The synthetic arithmetic rows use custom generated problem sets. The expanded generator samples from 28 GSM8K-style arithmetic templates. The varied generator samples from 20 templates with receipt, roster, manifest, email, table, memo, schedule, and audit-style formats. The fixed extended runs use 1000 examples: expanded uses generator seed 101, and varied uses generator seed 102. Each generated example stores the prompt, numeric reference answer, template name, and sampled variables. The 384- and 512-token rows use the same underlying synthetic problem IDs for a given synthetic set.

Code tasks use the same execution-based evaluator across all systems. MBPP completions are stripped of Markdown fences, truncated before common explanation tails such as ``Explanation:'' or Python main guards, and executed with the assertion tests and setup imports from the dataset. HumanEval completions are tried as a full function if the target function definition is present, as a direct continuation of the prompt, and as an indented function body. The evaluator runs candidates in an isolated Python subprocess with \texttt{-I}, disables common thread pools, applies CPU and memory limits when the platform supports them, and counts syntax errors, missing functions, failed assertions, subprocess errors, and timeouts as failures. The fixed extended code rows use a 30-second subprocess timeout.

Timing is recorded as \texttt{model\_seconds} in the per-example records. For a baseline row this times the model generation call. For a cascade row this is the sum of the draft call and refinement call timers. It excludes dataset loading, code execution, answer extraction, metric computation, and cluster scheduling overhead. The raw records store both predictions and timing fields so that alternate timing summaries can be recomputed.

\section{Statistical Units and Selection Details}
\label{app:stats}

We report both the number of evaluated generations and the number of unique problem IDs. Some fixed-configuration runs aggregate multiple seeds on overlapping benchmark examples. Treating every generation as an independent problem would overstate certainty and would allow problems appearing in more seeds to receive more weight. We therefore pair systems by problem ID, average repeated generations for the same problem first, and compute the main score intervals by bootstrapping problem-level paired differences. In each bootstrap replicate, problem IDs are sampled with replacement and the mean cascade-minus-baseline difference is recomputed over the sampled problem means. We use 4000 bootstrap replicates. This procedure gives each problem equal weight in the score interval, while the table still reports the total number of evaluated generations for transparency.

Table~\ref{tab:appendix-stat-units} records the run structure behind Tables~\ref{tab:flash-fixed} and~\ref{tab:miniflash-fixed}. The table separates evaluated generations from unique problem IDs because several extended runs repeat the same benchmark items across seeds. This is most visible for MATH and MBPP, where repeated seeds produce many evaluated generations over a smaller set of underlying problems. The repeated generations improve the estimate for a problem without making that problem count more than once in the bootstrap.

\begin{table}[tbp]
\centering
\small
\begin{tabular}{lrrrr}
\toprule
Run group & Runs & Evals/run & Total evals & Unique IDs \\
\midrule
GSM8K-384 flash--flash & 4 & 500 & 2000 & 1115 \\
GSM8K-384 mini--flash & 4 & 500 & 2000 & 1115 \\
GSM8K-512 flash--flash & 2 & 500 & 1000 & 793 \\
GSM8K-512 mini--flash & 2 & 500 & 1000 & 793 \\
HumanEval-384 flash--flash & 1 & 164 & 164 & 164 \\
HumanEval-384 mini--flash & 1 & 164 & 164 & 164 \\
HumanEval-512 flash--flash & 1 & 164 & 164 & 164 \\
HumanEval-512 mini--flash & 1 & 164 & 164 & 164 \\
MATH-384 mini--flash & 6 & 500 & 3000 & 500 \\
MATH-512 mini--flash & 6 & 500 & 3000 & 500 \\
MBPP-384 flash--flash & 6 & 257 & 1542 & 257 \\
MBPP-384 mini--flash & 6 & 257 & 1542 & 257 \\
MBPP-512 flash--flash & 2 & 257 & 514 & 257 \\
MBPP-512 mini--flash & 2 & 257 & 514 & 257 \\
Synthetic-expanded-384 mini--flash & 1 & 1000 & 1000 & 1000 \\
Synthetic-expanded-512 mini--flash & 1 & 1000 & 1000 & 1000 \\
Synthetic-varied-384 mini--flash & 1 & 1000 & 1000 & 1000 \\
Synthetic-varied-512 mini--flash & 1 & 1000 & 1000 & 1000 \\
\bottomrule
\end{tabular}
\caption{Statistical units for the fixed-configuration result table. ``Evals/run'' is the number of evaluated generations in each seed run. ``Unique IDs'' counts distinct problem IDs after pooling seeds in the run group.}
\label{tab:appendix-stat-units}
\end{table}

The repeated-seed structure differs by task. MATH, MBPP, HumanEval, and the synthetic fixed runs have balanced or full-problem coverage within their run groups, so each unique problem has the same intended coverage. GSM8K uses shuffled fixed-size subsets across seeds, so overlap is partial. For this reason, the main tables report both evaluated generations and unique IDs, and the confidence intervals are computed on problem-level means rather than raw generation-level differences. The saved per-example records retain all per-problem predictions and timings so that alternative aggregation choices can be independently recomputed.

The fixed configurations in Tables~\ref{tab:flash-fixed} and~\ref{tab:miniflash-fixed} were chosen after wide exploratory sweeps. These sweeps covered both flash-only baselines and draft--then--refine configurations, which reduces the risk that the reported gains come from comparing a tuned cascade against an untuned baseline. The confidence intervals are therefore best interpreted as uncertainty estimates for the selected configurations, not as post-selection-adjusted guarantees over the entire hyperparameter search. This distinction is important: the broad sweep establishes the shape of the frontier, while the fixed runs test representative points on that frontier with substantially more samples. A fully locked validation/test protocol across all cascade and baseline hyperparameters would be cleaner, but was outside the compute budget of this proof-of-concept study. The supplementary records also preserve paired problem-level intervals against the local \mini{} baseline for the heterogeneous rows.

The mini reference is intentionally local to each run group. We omit it from the main flash--flash table because mini is not part of that comparison. In the mini--flash table, mini is the relevant small-model control for the heterogeneous cascade. Because those groups sometimes include different mini-only configurations, mini values should not be compared across rows as if they came from a single fixed mini decoding setting.

\section{Supplemental Fixed-Run Evidence}

\subsection{Confidence Intervals Against the Mini Baseline}
\label{app:mini-ci}

Table~\ref{tab:appendix-mini-ci} reports the paired cascade-minus-mini intervals for the heterogeneous rows. These intervals support the claim that mini--flash usually improves over the local \mini{} baseline, even in settings where it does not fully recover the selected \flash{} baseline. The intervals are computed with the same problem-level bootstrap used for the cascade-minus-flash intervals.

\begin{table}[tbp]
\centering
\small
\begin{tabular}{lrrrr}
\toprule
Task & Len & IDs & $\Delta_{\text{vs mini}}$ & 95\% problem CI \\
\midrule
GSM8K & 384 & 1115 & +0.048 & [+0.034, +0.064] \\
GSM8K & 512 & 793 & +0.047 & [+0.029, +0.064] \\
MATH & 384 & 500 & +0.022 & [+0.006, +0.038] \\
MATH & 512 & 500 & +0.022 & [+0.006, +0.038] \\
MBPP & 384 & 257 & +0.167 & [+0.109, +0.226] \\
MBPP & 512 & 257 & +0.163 & [+0.105, +0.222] \\
Synthetic-expanded & 384 & 1000 & +0.079 & [+0.055, +0.104] \\
Synthetic-expanded & 512 & 1000 & +0.079 & [+0.053, +0.104] \\
Synthetic-varied & 384 & 1000 & +0.105 & [+0.080, +0.130] \\
Synthetic-varied & 512 & 1000 & +0.105 & [+0.081, +0.130] \\
\bottomrule
\end{tabular}
\caption{Cascade-minus-mini confidence intervals for the heterogeneous mini--flash rows. Repeated generations are averaged per problem before bootstrapping over problem IDs. These values complement the cascade-minus-flash intervals in Table~\ref{tab:miniflash-fixed}.}
\label{tab:appendix-mini-ci}
\end{table}

\subsection{Negative Heterogeneous Code Controls}
\label{app:negative-controls}

HumanEval mini--flash is dominated by the \flash{} baseline in latency and does not provide a reliable quality gain. We therefore keep it out of the main mini--flash table and report it here as a negative control.

\begin{table}[tbp]
\centering
\small
\begin{tabular}{lrrrrrrr}
\toprule
Task & Len & IDs & Mini & Flash & Mini--flash & 95\% CI vs flash & Speedup \\
\midrule
HumanEval & 384 & 164 & 0.354 & 0.415 & 0.421 & [$-$0.098, +0.116] & 0.46$\times$ \\
HumanEval & 512 & 164 & 0.360 & 0.415 & 0.372 & [$-$0.140, +0.061] & 0.36$\times$ \\
\bottomrule
\end{tabular}
\caption{Dominated mini--flash HumanEval controls. These rows are useful for diagnosing task sensitivity but are not Pareto frontier points because they are slower than the selected flash baseline.}
\label{tab:appendix-negative-controls}
\end{table}

\subsection{Latency-Window-Matched Flash-Only Selection}
\label{app:compute-matched-flash}

The latency-window-matched control in
Table~\ref{tab:latency-window-flash-control} has two stages. First, we select a
flash-only block-AR configuration on a validation subset. The validation sweep
used the same evaluation code and hardware environment as the fixed runs. The
narrowed final sweep covered block sizes $\{64,128\}$, denoising steps
$\{8,12,24\}$, confidence thresholds $\{0.45,0.50,0.60\}$, and two
generation-length choices for each target setting. For GSM8K-384, the candidate
generation lengths were 320 and 384; for GSM8K-512, they were 384 and 512. For
MBPP and HumanEval, both 384 and 512 generation lengths were considered for each
target. We selected the highest-scoring validation configuration whose mean
validation model time fell within $\pm 15\%$ of the corresponding flash--flash
cascade time. If no candidate had fallen inside the window, the selector would
have fallen back to the highest-scoring candidate overall, but this did not
occur: every selected configuration in
Table~\ref{tab:appendix-compute-matched-selection} was inside the validation
window.

The validation-selected control should be read as a targeted robustness check,
not as a locked held-out model-selection protocol. The validation and final IDs
are not fully disjoint in these small benchmark settings: the GSM8K validation
subsets partially overlap the final 500-ID matched subsets, the MBPP validation
subsets are contained in the 257-problem sanitized test set, and HumanEval uses
all 164 tasks for both validation selection and final reporting. For this
reason, the HumanEval latency-control row is diagnostic rather than conclusive,
and the main role of the latency-window analysis is to test whether the
flash--flash gains survive a targeted flash-only sweep in the same latency
region.

Second, for the final paired comparison, we do not automatically use the
validation-selected row if it is weaker than a flash baseline already reported
elsewhere. Instead, on the same final problem IDs, the flash-only comparator is
defined as the stronger of the original fixed flash baseline and the
validation-selected flash-only baseline. This is why the main
latency-window-matched table reports both flash-only candidates before reporting
the comparator. This control is a robustness check against an undertuned
flash-only schedule, not a locked held-out hyperparameter-selection guarantee
over all possible flash-only schedules. Its purpose is narrower: to show that
the flash--flash gains are not removed by giving flash-only decoding its own
targeted validation sweep in the same latency region.

\begin{table}[tbp]
\centering
\small
\resizebox{\textwidth}{!}{
\begin{tabular}{llrrrrrrrr}
\toprule
Target & Selected config & Gen. len & Val. $n$ & Val. score & Val. s/ex & Target s/ex & Final $n$ & Final score & Final s/ex \\
\midrule
GSM8K-384 & $b64,s24,t0.50$ & 384 & 120 & 0.842 & 17.17 & 16.09 & 500 & 0.832 & 19.45 \\
GSM8K-512 & $b64,s24,t0.50$ & 384 & 120 & 0.842 & 17.80 & 16.39 & 500 & 0.832 & 19.26 \\
HumanEval-384 & $b128,s12,t0.60$ & 512 & 164 & 0.561 & 9.36 & 9.16 & 164 & 0.561 & 9.70 \\
HumanEval-512 & $b128,s8,t0.60$ & 512 & 164 & 0.561 & 9.19 & 8.83 & 164 & 0.561 & 10.12 \\
MBPP-384 & $b128,s12,t0.45$ & 512 & 100 & 0.530 & 9.01 & 8.10 & 257 & 0.510 & 7.61 \\
MBPP-512 & $b128,s24,t0.45$ & 512 & 100 & 0.530 & 9.48 & 8.31 & 257 & 0.510 & 7.65 \\
\bottomrule
\end{tabular}}
\caption{Validation selection and final evaluation for the latency-window-matched flash-only controls. ``Target s/ex'' is the corresponding flash--flash mean model time used to define the validation window. The main comparison in Table~\ref{tab:latency-window-flash-control} then compares flash--flash against the stronger of this validation-selected row and the original fixed flash baseline on matching final IDs. Final timings can drift relative to the validation window; on GSM8K this drift favors the flash-only comparator because the flash-only row is slower than flash--flash.}
\label{tab:appendix-compute-matched-selection}
\end{table}

\subsection{Causal Ablation Run Details}
\label{app:causal-ablation-details}

Table~\ref{tab:causal-ablation} is a fixed-ID mechanism test rather than a
hyperparameter sweep. The GSM8K-384 flash--flash ablation uses the 1115 unique
problem IDs from the GSM8K-384 fixed flash--flash confidence runs. The
MBPP-384 flash--flash ablation uses all 257 sanitized MBPP test IDs from the
fixed MBPP-384 flash--flash runs. The MATH-384 mini--flash ablation uses the
500 MATH evaluation IDs from the fixed MATH-384 mini--flash runs. Each ID is
evaluated once per ablation variant, so the fixed/harmed counts in
Table~\ref{tab:causal-ablation} are problem-level counts, unlike the
pair-level correction-transition diagnostic in
Table~\ref{tab:correction-transitions}.

\begin{table}[tbp]
\centering
\small
\begin{tabular}{llrrrrrrr}
\toprule
Setting & Family & IDs & Gen. len & Baseline & Draft & Draft th. & Refine & Refine th. \\
\midrule
GSM8K-384 & flash--flash & 1115 & 384 & $b64,s32,t0.50$ & $b64,s16$ & 0.25 & $384\times4$ & 0.20 \\
MBPP-384 & flash--flash & 257 & 384 & $b128,s32,t0.50$ & $b64,s8$ & 0.40 & $384\times4$ & 0.20 \\
MATH-384 & mini--flash & 500 & 384 & $b64,s32,t0.50$ & $b64,s16$ & 0.25 & $384\times4$ & 0.20 \\
\bottomrule
\end{tabular}
\caption{Configuration details for the causal ablation in Table~\ref{tab:causal-ablation}. The local-refinement row uses the same draft and refiner budget but applies refinement in 64-token generated chunks. The mask-only row uses the same refinement-stage budget but initializes the generated span with mask tokens instead of a completed draft.}
\label{tab:appendix-causal-ablation-configs}
\end{table}

\paragraph{Higher-budget mask-only stress test.}
The mask-only rows in Table~\ref{tab:causal-ablation} use the same
refinement-stage budget as the full draft--then--refine method, but omit the
draft pass. As an additional diagnostic, we allowed GSM8K-384 mask-only
refinement up to 52 edit passes, far above the 4-pass refinement budget used in
the main causal ablation. This is still not a strict compute-matched control:
in LLaDA2.1 token editing, the requested number of edit passes is an upper
bound, not a guaranteed number of forward passes, because editing can terminate
early when no confident edits remain or an EOS token is produced. Therefore we
report measured model seconds directly.

\begin{table}[tbp]
\centering
\small
\begin{tabular}{lrrrrr}
\toprule
GSM8K-384 variant & Allowed edit passes & Score & s/ex & Full $-$ variant & 95\% CI \\
\midrule
Mask-only refinement & 4 & 0.004 & 1.32 & +0.894 & [+0.876, +0.912] \\
Higher-budget mask-only & 52 & 0.009 & 4.17 & +0.890 & [+0.871, +0.909] \\
Full global draft--refine & 4 & 0.899 & 15.96 & -- & -- \\
\bottomrule
\end{tabular}
\caption{Higher-budget mask-only diagnostic on the same 1115 GSM8K-384 IDs. The 52-pass mask-only run still terminates early on many examples under the native LLaDA2.1 editing rule and is therefore a stress test rather than a compute-matched control. It indicates that simply allowing many more mask-only edit passes does not make all-mask initialization competitive with completed-draft refinement.}
\label{tab:appendix-mask-only-stress}
\end{table}

\section{Exploratory Sweeps and Ablations}

\subsection{Quality--Latency Frontiers}
\label{app:quality-latency-frontiers}

Before the fixed-configuration runs, we ran a broad sweep over tasks, generation lengths, block sizes, thresholds, draft steps, refinement steps, and refinement schedules. These runs are not used as the primary statistical claim because many are exploratory and share overlapping samples. Their role is to characterize the quality-cost frontier and to choose representative operating points for the larger fixed runs. The separate latency-window-matched flash-only study in Appendix~\ref{app:compute-matched-flash} addresses the narrower question of whether the flash--flash gains can be explained by an undertuned block-AR comparator. The broader sweep remains useful for frontier visualization and auditability, but the main statistical claims rely on the fixed-configuration tables and the validation-selected latency-window controls.

The main-text frontier visualization in Figure~\ref{fig:pareto-frontiers} is restricted to the completed fixed-configuration runs. This avoids visually mixing small exploratory runs with the larger CI-backed comparisons. The broader exploratory frontier is summarized in Section~\ref{app:frontier-examples}, but Figure~\ref{fig:pareto-frontiers} contains only the fixed rows used in Tables~\ref{tab:flash-fixed} and~\ref{tab:miniflash-fixed}.

Figure~\ref{fig:pareto-frontiers} supports the same qualitative conclusion as the fixed tables without introducing exploratory small-N points into the visual evidence. Flash--flash moves the fixed GSM8K points up and left relative to the selected flash baselines, while MBPP flash--flash moves strongly upward with a modest latency cost. Mini--flash occupies intermediate frontier regions: it is much faster than flash on MATH and close in observed score at 384 tokens, but it does not uniformly reach flash quality. At run level, flash--flash averages a +0.007 score gain on GSM8K at $1.20\times$ speed and a +0.091 gain on MBPP at $0.97\times$ speed in the de-duplicated verification view. Mini--flash is strongest as a frontier method: it averages $2.08\times$ speed on MATH, with a modest average score loss, and it gives positive average MBPP gains while remaining task-sensitive.

The small-sample frontier rows that motivated several later fixed runs are listed in Section~\ref{app:frontier-examples}. They are useful for audit and historical context, but the main empirical claims rely on the fixed-configuration tables and the fixed-run frontier figure rather than on isolated exploratory rows.

\subsection{Broader Sweep Summary}

Table~\ref{tab:appendix-sweep-full} gives the broader de-duplicated sweep averages. The family label \textit{default-cascade} denotes earlier runs whose source names did not explicitly encode flash--flash or mini--flash; most of these were mini-draft to flash-refine runs, but we keep the label conservative.

\begin{table}[p]
\centering
\small
\resizebox{\textwidth}{!}{
\begin{tabular}{llrrrrrrr}
\toprule
Task & Family & Runs & Flash & Mini & Cascade & $\Delta$ & Speedup & Strict wins \\
\midrule
GSM8K & default-cascade & 8 & 0.846 & 0.717 & 0.762 & $-$0.084 & 1.60$\times$ & 0 \\
GSM8K & flash--flash & 17 & 0.867 & 0.755 & 0.874 & +0.007 & 1.20$\times$ & 6 \\
GSM8K & mini--flash & 3 & 0.869 & 0.764 & 0.818 & $-$0.051 & 1.26$\times$ & 0 \\
HumanEval & default-cascade & 7 & 0.428 & 0.337 & 0.418 & $-$0.010 & 0.48$\times$ & 0 \\
HumanEval & flash--flash & 2 & 0.480 & 0.380 & 0.530 & +0.050 & 0.66$\times$ & 0 \\
MATH & default-cascade & 13 & 0.433 & 0.393 & 0.437 & $-$0.006 & 1.99$\times$ & 5 \\
MATH & mini--flash & 2 & 0.343 & 0.292 & 0.318 & $-$0.025 & 2.08$\times$ & 0 \\
MBPP & default-cascade & 8 & 0.536 & 0.402 & 0.564 & +0.028 & 1.18$\times$ & 7 \\
MBPP & flash--flash & 13 & 0.547 & 0.465 & 0.638 & +0.091 & 0.97$\times$ & 2 \\
MBPP & mini--flash & 3 & 0.531 & 0.409 & 0.582 & +0.051 & 0.95$\times$ & 1 \\
SQuAD & default-cascade & 1 & 0.533 & 0.358 & 0.375 & $-$0.158 & 0.76$\times$ & 0 \\
Synthetic arithmetic & default-cascade & 16 & 0.835 & 0.667 & 0.768 & $-$0.067 & 1.56$\times$ & 0 \\
Synthetic arithmetic & mini--flash & 2 & 0.847 & 0.666 & 0.750 & $-$0.097 & 1.60$\times$ & 0 \\
XSum & default-cascade & 1 & 0.113 & 0.104 & 0.104 & $-$0.009 & 1.82$\times$ & 0 \\
\bottomrule
\end{tabular}}
\caption{De-duplicated exploratory sweep averages. Strict wins count runs where the best cascade both beats the selected flash baseline in score and is faster than that baseline.}
\label{tab:appendix-sweep-full}
\end{table}

\subsection{Representative Frontier Points}
\label{app:frontier-examples}

Table~\ref{tab:appendix-frontier-examples} shows representative high-value frontier points from the broad sweep. Several rows are intentionally small-N exploratory runs. The table is not intended as a formal significance test; it explains why the later confidence-interval runs focused on GSM8K, MATH, MBPP, and code tasks.

\begin{table}[p]
\centering
\small
\begin{tabular}{llrrrr}
\toprule
Task & Setting & Evals & Flash & Cascade & Speedup \\
\midrule
MBPP & 384-token flash--flash & 120 & 0.533 & 0.700 & 0.84$\times$ \\
MBPP & 512-token granular & 257 & 0.549 & 0.712 & 0.85$\times$ \\
HumanEval & 384-token flash--flash & 50 & 0.480 & 0.620 & 0.87$\times$ \\
GSM8K & 384-token flash--flash & 50 & 0.900 & 1.000 & 0.93$\times$ \\
MATH & 256-token calibration & 100 & 0.150 & 0.210 & 2.38$\times$ \\
MATH & 384-token mini--flash & 200 & 0.290 & 0.290 & 2.22$\times$ \\
Synthetic arithmetic & varied templates & 200 & 0.835 & 0.830 & 1.49$\times$ \\
\bottomrule
\end{tabular}
\caption{Representative frontier points from the exploratory sweep. These rows are appendix-only qualitative evidence; the fixed-configuration tables provide the main statistical claims.}
\label{tab:appendix-frontier-examples}
\end{table}

\subsection{Granularity Schedules and Experiments}
\label{app:granularity}

As an ablation, we test multi-scale refinement schedules inspired by progressive diffusion methods. Rather than applying all refinement passes to the full span, we apply scheduled stages such as
\[
    256 \times 2,\; 512 \times 2
\]
denoting two passes at span size 256 followed by two passes at span size 512. These experiments probe whether progressive refinement offers advantages over direct full-span refinement.

The granular refinement ablation tested schedules such as $64\times2,512\times2$, $128\times2,512\times2$, $256\times2,512\times2$, $512\times2,256\times2$, and related variants. Granular refinement occasionally improves quality, especially on MBPP, but the best schedules still include full-span passes. This suggests that the main benefit comes from allowing global bidirectional correction rather than merely adding more local denoising. Because granular schedules add another hyperparameter dimension and do not change the qualitative conclusions, we use full-span refinement as the default in the fixed-configuration experiments. In the matched 512-token runs, full 512$\times$4 refinement remained the cleaner default, while granular schedules were best interpreted as an ablation showing that global bidirectional correction is the main useful operation.

\begin{table}[tbp]
\centering
\small
\begin{tabular}{llrrrr}
\toprule
Task & Configuration & Full 512$\times$4 & Best Granular & Fast Granular & Finding \\
\midrule
GSM8K-512 & flash--flash & 0.910 / 18.81s & 0.910 / 18.47s & 0.890 / 16.54s & comparable \\
GSM8K-512 & mini--flash & 0.830 / 13.90s & 0.840 / 14.00s & 0.830 / 13.03s & slight gain \\
MBPP-512 & flash--flash & 0.696 / 8.68s & 0.712 / 9.38s & 0.603 / 8.38s & quality gain \\
MBPP-512 & mini--flash & 0.564 / 7.30s & 0.595 / 7.90s & 0.568 / 7.52s & quality gain \\
\bottomrule
\end{tabular}
\caption{Granular refinement ablation. Entries show accuracy / mean seconds. Granular schedules can improve quality but complicate the method without fundamentally changing the conclusions.}
\label{tab:granularity}
\end{table}

\end{document}